%% file: main.tex
\documentclass[10pt,twocolumn,letterpaper]{article}
\usepackage[accsupp]{axessibility}  %
\usepackage{iccv}
\usepackage{times}
\usepackage{epsfig}
\usepackage{graphicx}
\usepackage{amsmath}
\usepackage{amssymb}
\input{assets/macros}

\usepackage{hyperref}
\hypersetup{
colorlinks=true,
linkcolor=black,
citecolor=blue,
filecolor=magenta,
urlcolor=blue
}

\iccvfinalcopy %

\def\iccvPaperID{1900} %

\ificcvfinal\fi

\begin{document}

\title{Perpetual Humanoid Control for Real-time Simulated Avatars}

\author{
Zhengyi Luo$^{1,2}$
\quad
Jinkun Cao$^{2}$
\quad
Alexander Winkler$^{1}$
\quad
Kris Kitani$^{1,2}$
\quad
Weipeng Xu$^{1}$ \\
$^{1}$Reality Labs Research, Meta; $^{2}$Carnegie Mellon University\\
{\tt\small \href{https://zhengyiluo.github.io/PHC/}{https://zhengyiluo.github.io/PHC/}  }}
\input{assets/main/figures/teaser}
\maketitle

\ificcvfinal\thispagestyle{empty}\fi

\begin{abstract}
    We present a physics-based humanoid controller that achieves high-fidelity motion imitation and fault-tolerant behavior in the presence of noisy input (\eg pose estimates from video or generated from language) and unexpected falls. Our controller scales up to learning ten thousand motion clips without using any external stabilizing forces and learns to naturally recover from fail-state. Given reference motion, our controller can perpetually control simulated avatars without requiring resets. At its core, we propose the progressive multiplicative control policy (PMCP), which dynamically allocates new network capacity to learn harder and harder motion sequences. PMCP allows efficient scaling for learning from large-scale motion databases and adding new tasks, such as fail-state recovery, without catastrophic forgetting. We demonstrate the effectiveness of our controller by using it to imitate noisy poses from video-based pose estimators and language-based motion generators in a live and real-time multi-person avatar use case.

\end{abstract}

\etocdepthtag.toc{mtchapter}
\etocsettagdepth{mtchapter}{subsection}
\etocsettagdepth{mtappendix}{none}

\vspace{-5mm}

\section{Introduction}

Physics-based motion imitation has captured the imagination of vision and graphics communities due to its potential for creating realistic human motion, enabling plausible environmental interactions, and advancing virtual avatar technologies of the future. However, controlling high-degree-of-freedom (DOF) humanoids in simulation presents significant challenges, as they can fall, trip, or deviate from their reference motions, and struggle to recover. For example, controlling simulated humanoids using poses estimated from noisy video observations can often lead humanoids to fall to the ground\cite{Yuan2018-ft, Yuan2019-qp, Luo2021-gu, Luo2022-qc}. These limitations prevent the widespread adoption of physics-based methods, as current control policies cannot handle noisy observations such as video or language. 

In order to apply physically simulated humanoids for avatars, the first major challenge is learning a motion imitator (controller) that can faithfully reproduce human-like motion with a high success rate. While reinforcement learning (RL)-based imitation policies have shown promising results, successfully imitating motion from a large dataset, such as AMASS (ten thousand clips, 40 hours of motion), with a \textit{single} policy has yet to be achieved. Attempts to use larger or a mixture of expert policies have been met with some success \cite{Wang2020-qi, Won2020-lb}, although they have not yet scaled to the largest dataset. Therefore, researchers have resorted to using external forces to help stabilize the humanoid. Residual force control (RFC) \cite{Yuan2020-fp} has helped to create motion imitators that can mimic up to 97\% of the AMASS dataset \cite{Luo2021-gu}, and has seen successful applications in human pose estimation from video\cite{Yuan2021-rl, Luo2022-ux, Gong2022-sv} and language-based motion generation \cite{Yuan2022-re}. However, the external force compromises physical realism by acting as a ``hand of God" that puppets the humanoid, leading to artifacts such as flying and floating. One might argue that, with RFC, the realism of simulation is compromised, as the model can freely apply a non-physical force on the humanoid.

Another important aspect of controlling simulated humanoids is how to handle noisy input and failure cases. In this work, we consider human poses estimated from video or language input. Especially with respect to video input, artifacts such as floating \cite{Yuan2022-re}, foot sliding \cite{Zou2020-bl}, and physically impossible poses are prevalent in popular pose estimation methods due to occlusion, challenging view point and lighting, fast motions \etc. To handle these cases, most physics-based methods resort to resetting the humanoid when a failure condition is triggered \cite{Luo2022-qc, Luo2021-gu, Yuan2019-qp}. However, resetting successfully requires a high-quality reference pose, which is often difficult to obtain due to the noisy nature of the pose estimates, leading to a vicious cycle of falling and resetting to unreliable poses. Thus, it is important to have a controller that can gracefully handle unexpected falls and noisy input, naturally recover from fail-state, and resume imitation.

In this work, our aim is to create a humanoid controller specifically designed to control real-time virtual avatars, where video observations of a human user are used to control the avatar. We design the Perpetual Humanoid Controller (PHC), a \textit{single} policy that achieves a high success rate on motion imitation \textbf{and} can recover from fail-state naturally. We propose a progressive multiplicative control policy (PMCP) to learn from motion sequences in the entire AMASS dataset without suffering catastrophic forgetting. By treating harder and harder motion sequences as a different ``task" and gradually allocating new network capacity to learn, PMCP retains its ability to imitate easier motion clips when learning harder ones. PMCP also allows the controller to learn fail-state recovery tasks \textit{without compromising} its motion imitation capabilities. Additionally, we adopt Adversarial Motion Prior (AMP)\cite{Peng2021-xu} throughout our pipeline and ensure natural and human-like behavior during fail-state recovery. Furthermore, while most motion imitation methods require both estimates of link position and rotation as input, we show that we can design controllers that require only the link positions. This input can be generated more easily by vision-based 3D keypoint estimators or 3D pose estimates from VR controllers.

To summarize, our contributions are as follows: (1) we propose a Perpetual Humanoid Controller that can successfully imitate 98.9\% of the AMASS dataset without applying any external forces; (2) we propose the progressive multiplicative control policy to learn from a large motion dataset without catastrophic forgetting and unlock additional capabilities such as fail-state recovery; (3) our controller is task-agnostic and is compatible with off-the-shelf video-based pose estimators as a drop-in solution. We demonstrate the capabilities of our controller by evaluating on both Motion Capture (MoCap) and estimated motion from videos. We also show a live (30 fps) demo of driving perpetually simulated avatars using a webcam video as input.

\section{Related Works}
\paragraph{Physics-based Motion Imitation} Governed by the laws of physics, simulated characters \cite{Peng2017-il, Peng2018-fu, Peng2019-kf, Peng2021-xu, Peng2022-vr, Chentanez2018-cw, Wang2020-qi, Yuan2020-fp, Merel2020-qm, Hasenclever_undated-cs, Bergamin2019-ak, Fussell2021-jh, Winkler2022-bv, Gong2022-sv} have the distinct advantage of creating natural human motion, human-to-human interaction \cite{Liu2021-iz, Won2021-sn}, and human-object interactions \cite{Merel2020-qm, Peng2022-vr}. Since most modern physics simulators are not differentiable, training these simulated agents requires RL, which is time-consuming \& costly. As a result, most of the work focuses on small-scale use cases such as interactive control based on user input \cite{Wang2020-qi, Bergamin2019-ak, Peng2021-xu,Peng2022-vr}, playing sports \cite{Won2021-sn, Liu2021-iz, Merel2020-qm}, or other modular tasks (reaching goals \cite{Won2022-jy}, dribbling \cite{Peng2021-xu}, moving around \cite{Peng2017-il}, \etc). On the other hand, imitating large-scale motion datasets is a challenging yet fundamental task, as an agent that can imitate reference motion can be easily paired with a motion generator to achieve different tasks. From learning to imitate a single clip \cite{Peng2018-fu} to datasets \cite{Won2020-lb, Wang2020-qi, Chentanez2018-cw, Wagener2022-aj}, motion imitators have demonstrated their impressive ability to imitate reference motion, but are often limited to imitating high-quality MoCap data. Among them, ScaDiver \cite{Won2020-lb} uses a mixture of expert policy to scale up to the CMU MoCap dataset and achieves a success rate of around 80\% measured by time to failure. Unicon\cite{Wang2020-qi} shows qualitative results in imitation and transfer, but does not quantify the imitator's ability to imitate clips from datasets. MoCapAct\cite{Wagener2022-aj} first learns single-clip experts on the CMU MoCap dataset, and distills them into a single that achieves around 80\% of the experts' performance. The effort closest to ours is UHC \cite{Luo2021-gu}, which successfully imitates 97\% of the AMASS dataset. However, UHC uses residual force control \cite{Yuan2019-qp}, which applies a non-physical force at the root of the humanoid to help balance. Although effective in preventing the humanoid from falling, RFC reduces physical realism and creates artifacts such as floating and swinging, especially when motion sequences become challenging \cite{Luo2021-gu, Luo2022-ux}. Compared to UHC, our controller does not utilize any external force.

\paragraph{Fail-state Recovery for Simulated Characters} As simulated characters can easily fall when losing balance, many approaches \cite{Shimada2020-wv, Yuan2019-qp, Peng2022-vr, Tao2022-ed, Chentanez2018-cw} have been proposed to help recovery. PhysCap \cite{Shimada2020-wv} uses a floating-base humanoid that does not require balancing. This compromises physical realism, as the humanoid is no longer properly simulated. Egopose \cite{Yuan2019-qp} designs a fail-safe mechanism to reset the humanoid to the kinematic pose when it is about to fall, leading to potential teleport behavior in which the humanoid keeps resetting to unreliable kinematic poses. NeruoMoCon \cite{Huang2022-bc} utilizes sampling-based control and reruns the sampling process if the humanoid falls. Although effective, this approach does not guarantee success and prohibits real-time use cases. Another natural approach is to use an additional recovery policy \cite{Chentanez2018-cw} when the humanoid has deviated from the reference motion. However, since such a recovery policy no longer has access to the reference motion, it produces unnatural behavior, such as high-frequency jitters. To combat this, ASE \cite{Peng2022-vr} demonstrates the ability to rise naturally from the ground for a sword-swinging policy. While impressive, in motion imitation the policy not only needs to get up from the ground, but also goes back to tracking the reference motion. In this work, we propose a comprehensive solution to the fail-state recovery problem in motion imitation: our PHC can rise from fallen state and naturally walks back to the reference motion and resume imitation.

\paragraph{Progressive Reinforcement Learning} When learning from data containing diverse patterns, catastrophic forgetting~\cite{French2002-is, McCloskey1989-vo} is observed when attempting to perform multi-task or transfer learning by fine-tuning. Various approaches \cite{De_Lange2022-an, Jia2022-jm, Kirkpatrick2017-ee} have been proposed to combat this phenomenon, such as regularizing the weights of the network \cite{Kirkpatrick2017-ee}, learning multiple experts \cite{Jia2022-jm}, or increasing the capacity using a mixture of experts \cite{Zhou2022-iw, Shazeer2017-hs, Won2020-lb} or multiplicative control \cite{Peng2019-kf}. A paradigm has been studied in transfer learning and domain adaption as progressive learning~\cite{Chen2018-fs,Cao2019-ip} or curriculum learning~\cite{Bengio2009-kw}. Recently, progressive reinforcement learning~\cite{Berseth2018-sv} has been proposed to distill skills from multiple expert policies. It aims to find a policy that best matches the action distribution of experts instead of finding an optimal mix of experts.  Progressive Neural Networks (PNN)~\cite{Rusu2016-go} proposes to avoid catastrophic forgetting by freezing the weights of the previously learned subnetworks and initializing additional subnetworks to learn new tasks. The experiences from previous subnetworks are forwarded through lateral connections. PNN requires manually choosing which subnetwork to use based on the task, preventing it from being used in motion imitation since reference motion does not have the concept of task labels.

\section{Method}
We define the reference pose as $\refp \triangleq (\refr, \reft)$, consisting of 3D joint rotation $\refr \in \reals^{J \times 6}$ and position $\reft \in \reals^{J \times 3}$ of all $J$ links on the humanoid (we use the 6 DoF rotation representation \cite{Zhou2019-lj}). From reference poses $\refps$, one can compute the reference velocities $\refvs$ through finite difference, where $\refv \triangleq (\refav, \reflv)$ consist of angular $\refav \in \reals^{J \times 3}$ and linear velocities $\reflv \in \reals^{J \times 3}$. We differentiate rotation-based and keypoint-based motion imitation by input: rotation-based imitation relies on reference poses $\refps$ (both rotation and keypoints), while keypoint-based imitation only requires 3D keypoints $\refts$. As a notation convention, we use $\widetilde{\cdot}$ to represent kinematic quantities (without physics simulation) from pose estimator/keypoint detectors, $\widehat{\cdot}$ to denote ground truth quantities from Motion Capture (MoCap), and normal symbols without accents for values from the physics simulation. We use ``imitate", ``track", and ``mimic" reference motion interchangeably. In Sec.\ref{sec:goal-rl}, we first set up the preliminary of our main framework. Sec.\ref{sec:pmcp} describes our progressive multiplicative control policy to learn to imitate a large dataset of human motion and recover from fail-states. Finally, in Sec.\ref{sec:real-time}, we briefly describe how we connect our task-agnostic controller to off-the-shelf video pose estimators and generators for real-time use cases.

\subsection{Goal Conditioned Motion Imitation with Adversarial Motion Prior}
\label{sec:goal-rl}
Our controller follows the general framework of goal-conditioned RL (Fig.\ref{fig:arch}), where a goal-conditioned policy $\policy$ is tasked to imitate reference motion $\bs{\hat{q}_{1:t}}$ or keypoints $\refts$. Similar to prior work \cite{Luo2021-gu, Peng2018-fu}, we formulate the task as a Markov Decision Process (MDP) defined by the tuple ${\mathcal M}=\langle \mathcal{\bs S}, \mathcal{ \bs A}, \mathcal{ \bs T}, \rewardfunc, \gamma\rangle$ of states, actions, transition dynamics, reward function, and discount factor. The physics simulation determines state $\state \in \mathcal{ \bs S}$ and transition dynamics $\mathcal{ \bs T}$ while our policy $\policy$ computes per-step action $\bs{a}_t \in \mathcal{ \bs A}$. Based on the simulation state $\state$ and reference motion $\refp$, the reward function $\rewardfunc$ computes a reward $r_t = \rewardfunc(\state, \refp)$  as the learning signal for our policy. The policy's goal is to maximize the discounted reward $\mathbb{E}\left[\sum_{t=1}^{T} \gamma^{t-1} r_{t}\right]$, and we use the proximal policy gradient (PPO) \cite{Schulman2017-ft} to learn $\policy$.

\input{assets/main/figures/train}

\input{assets/main/figures/arch}

\paragraph{State} The simulation state $\state \triangleq (\selfstate, \bs{s}^{\text{g}}_t)$ consists of humanoid proprioception $\selfstate$ and the goal state $\goalstate$. Proprioception $\selfstate \triangleq (\simp, \simv, \bs\beta)$ contains the 3D body pose $\simp$, velocity $\simv$, and (optionally) body shapes $\bs \beta$. When trained with different body shapes, $\bs \beta$ contains information about the length of the limb of each body link \cite{Luo2022-qc}. For rotation-based motion imitation, the goal state $\goalstate$ is defined as the difference between the next time step reference quantitives and their simulated counterpart: $$\goalstaterot \triangleq (  \refrn \ominus \simr, \reftn -  \simt, \bs{\hat{v}}_{t+1} -  \bs{v}_t, \bs{\hat{\omega}}_t -  \bs{\omega}_t, \refrn, \reftn)$$ where $\ominus $ calculates the rotation difference. For keypoint-only imtiation, the goal state becomes $$\goalstatekp \triangleq ( \reftn -  \simt, \bs{\hat{v}}_{t+1} -  \simlv,  \reftn).$$ All of the above quantities in $\goalstate$ and $\selfstate$ are normalized with respect to the humanoid's current facing direction and root position \cite{Won2022-jy, Luo2021-gu}. 

\paragraph{Reward} Unlike prior motion tracking policies that only use a motion imitation reward, we use the recently proposed Adversarial Motion Prior \cite{Peng2021-xu} and include a discriminator reward term throughout our framework.  Including the discriminator term helps our controller produce stable and natural motion and is especially crucial in learning natural fail-state recovery behaviors. Specifically, our reward is defined as the sum of a task reward $r^{\text{g}}_t$, a style reward $r^{\text{amp}}_t$, and an additional energy penalty $r^{\text{energy}}_t$ \cite{Peng2018-fu}: 
\begin{equation} 
    r^{\text{}}_t = 0.5 r^{\text{g}}_t + 0.5 r^{\text{amp}}_t + r^{\text{energy}}_t. 
\end{equation}
For the discriminator, we use the same observations, loss formulation, and gradient penalty as AMP \cite{Peng2021-xu}. The energy penalty is expressed as $ -0.0005 \cdot \sum_{j \in \text { joints }}\left|\bs{\mu_j} \bs{\omega_j}\right|^2 $ where $\bs{\mu_j}$ and $\bs{\omega_j}$ correspond to the joint torque and the joint angular velocity, respectively. The energy penalty \cite{Fu2022-ky} regulates the policy and prevents high-frequency jitter of the foot that can manifest in a policy trained without external force (see Sec.\ref{sec:imitation}). The task reward is defined based on the current training objective, which can be chosen by switching the reward function for motion imitation $\rewardfuncimitation$ and fail-state recovery $\rewardfuncfailrec$. For motion tracking, we use:

\small
\begin{equation}
    \begin{aligned}
    & r^{\text{g-imitation}}_t  =  \rewardfuncimitation(\state, \refp)  =  w_{\text{jp}} e^{-100 \| \bs{\hat p}_t - \bs{p}_t \|} \\ & + w_{\text{jr}} e^{-10\| \bs{\hat q}_t \ominus \bs{q}_t \|}  + w_{\text{jv}} e^{-0.1\| \bs{\hat v}_t - \bs{v}_t \|} + w_{\text{j}\omega} e^{-0.1\| \bs{\hat \omega}_t - \bs{\omega}_t \|} 
    \end{aligned}
\end{equation}
where we measure the difference between the translation, rotation, linear velocity, and angular velocity of the rigid body for all links in the humanoid. For fail-state recovery, we define the reward $r^{\text{g-recover}}_t $ in Eq.\ref{eqn:fail_recovery}. 
 
\paragraph{Action} We use a proportional derivative (PD) controller at each DoF of the humanoid and the action $\bs a_t$ specifies the PD target. With the target joint set as $\bs{q}_t^d = \action$, the torque applied at each joint is $\boldsymbol{\tau}^i=\boldsymbol{k}^p \circ\left(\action-\boldsymbol{q}_t\right)-\boldsymbol{k}^d \circ \dot{\boldsymbol{q}}_t $. Notice that this is different from the residual action representation \cite{Yuan2020-fp, Luo2021-gu, Park2019-ux} used in prior motion imitation methods, where the action is added to the reference pose: $\bs q^d_t = \hat{\bs q}_t + \bs{a}_t$ to speed up training. As our PHC needs to remain robust to noisy and ill-posed reference motion, we remove such a dependency on reference motion in our action space. We do not use any external forces \cite{Yuan2020-fp} or meta-PD control\cite{Yuan2021-rl}.

\paragraph{Control Policy and Discriminator} Our control policy $\policy$$(\bs{a}_t | \state) = \mathcal{N} (\mu (\state), \sigma)$ represents a Gaussian distribution with fixed diagonal covariance. The AMP discriminator $\disc ( \bs{s}^{\text{p}}_{t-10:t})$ computes a real and fake value based on the current prioproception of the humanoid. All of our networks (discriminator, primitive, value function, and discriminator) are two-layer multilayer perceptrons (MLP) with dimensions [1024, 512].

\paragraph{Humanoid}
Our humanoid controller can support any human kinematic structure, and we use the SMPL \cite{Loper2015-ey} kinematic structure following prior arts \cite{Yuan2021-rl, Luo2021-gu,  Luo2022-ux}. The SMPL body contains 24 rigid bodies, of which 23 are actuated, resulting in an action space of $\bs{a}_t \in \reals^{23 \times 3}$. The body proportion can vary based on a body shape parameter $
\beta \in \reals^{10}$.

\paragraph{Initialization and Relaxed Early Termination} We use reference state initialization (RSI) \cite{Peng2018-fu} during training and randomly select a starting point for a motion clip for imitation. For early termination, we follow UHC \cite{Luo2021-gu} and terminate the episode when the joints are more than 0.5 meters globally on average from the reference motion. Unlike UHC, we remove the ankle and toe joints from the termination condition. As observed by RFC \cite{Yuan2020-fp}, there exists a dynamics mismatch between simulated humanoids and real humans, especially since the real human foot is multisegment \cite{Park2018-ks}. Thus, it is not possible for the simulated humanoid to have the exact same foot movement as MoCap, and blindly following the reference foot movement may lead to the humanoid losing balance. Thus, we propose Relaxed Early Termination (RET), which allows the humanoid's ankle and toes to slightly deviate from the MoCap motion to remain balanced. Notice that the humanoid still receives imitation and discriminator rewards for these body parts, which prevents these joints from moving in a nonhuman manner. We show that though this is a small detail, it is conducive to achieving a good motion imitation success rate. 

\paragraph{Hard Negative Mining} When learning from a large motion dataset, it is essential to train on harder sequences in the later stages of training to gather more informative experiences. We use a similar hard negative mining procedure as in UHC \cite{Luo2021-gu} and define hard sequences by whether or not our controller can successfully imitate this sequence. From a motion dataset $\motiondata$, we find hard sequences $\hardmotiondata \subseteq \motiondata$ by evaluating our model over the entire dataset and choosing sequences that our policy fails to imitate. 
\vspace{-5mm}
\subsection{Progressive Multiplicative Control Policy}

\label{sec:pmcp}
 As training continues, we notice that the performance of the model plateaus as it forgets older sequences when learning new ones. Hard negative mining alleviates the problem to a certain extent, yet suffers from the same issue. Introducing new tasks, such as fail-state recovery, may further degrade imitation performance due to catastrophic forgetting. These effects are more concretely categorized in the Appendix (App. C). Thus, we propose a progressive multiplicative control policy (PMCP), which allocates new subnetworks (primitives $\prim$) to learn harder sequences.

\paragraph{Progressive Neural Networks (PNN)} A PNN \cite{Rusu2016-go} starts with a single primitive network $\prim^{(1)}$ trained on the full dataset $\motiondata$. Once $\prim^{(1)}$ is trained to convergence on the entire motion dataset $\motiondata$ using the imitation task, we create a subset of hard motions by evaluating $\prim^{(1)}$ on $\motiondata$. We define convergence as the success rate on $\hardmotiondata^{(k)}$ no longer increases. The sequences that $\prim^{(1)}$ fails on is formed as $\hardmotiondata^{(1)}$. We then freeze the parameters of $\prim^{(1)}$ and create a new primitive $\prim^{(2)}$ (randomly initialized) along with lateral connections that connect each layer of $\prim^{(1)}$ to $\prim^{(2)}$. For more information about PNN, please refer to our supplementary material. During training, we construct each $\hardmotiondata^{(k)}$ by selecting the failed sequences from the previous step $\hardmotiondata^{(k-1)}$, resulting in a smaller and smaller hard subset: $\hardmotiondata^{(k)} \subseteq \hardmotiondata^{(k-1)}$. In this way, we ensure that each newly initiated primitive $\prim^{(k)}$  is responsible for learning a new and harder subset of motion sequences, as can be seen in Fig.\ref{fig:train}.  Notice that this is different from hard-negative mining in UHC \cite{Luo2021-gu}, as we initialize a new primitive $\prim^{(k+1)}$ to train. Since the original PNN is proposed to solve completely new tasks (such as different Atari games), a lateral connection mechanism is proposed to allow later tasks to choose between reuse, modify, or discard prior experiences. However, mimicking human motion is highly correlated, where fitting to harder sequences $\hardmotiondata^{(k)}$ can effectively draw experiences from previous motor control experiences. Thus, we also consider a variant of PNN where there are \textbf{no lateral} connections, but the new primitives are initialized from the weights of the prior layer. This weight sharing scheme is similar to fine-tuning on the harder motion sequences using a new primitive $\prim^{(k+1)}$ and preserve $\prim^{(k)}$'s ability to imitate learned sequences. 

\input{assets/main/tables/algo_train}

\input{assets/main/figures/qual}
\paragraph{Fail-state Recovery}
In addition to learning harder sequences, we also learn new tasks, such as recovering from fail-state. We define three types of fail-state: 1) fallen on the ground; 2) faraway from the reference motion ($> 0.5m$); 3) their combination: fallen and faraway. In these situations, the humanoid should get up from the ground, approach the reference motion in a natural way, and resume motion imitation. For this new task, we initialize a primitive $\prim^{(F)}$ at the end of the primitive stack. $\prim^{(F)}$ shares the same input and output space as $\prim^{(1)} \cdots \prim^{(k)}$, but since the reference motion does not provide useful information about fail-state recovery (the humanoid should not attempt to imitate the reference motion when lying on the ground), we modify the state space during fail-state recovery to remove all information about the reference motion except the root.  For the reference joint rotation $\refr = [\bs{\hat{\theta}}_t^{0}, \bs{\hat{\theta}}_t^{1}, \cdots \bs{\hat{\theta}}_t^{J}]$ where $\bs{\hat{\theta}}_t^{i}$ corresponds to the $i^{\text{th}}$ joint, we construct $\refr^{\prime} = [\bs{\hat{\theta}}_t^{0}, \bs{{\theta}}_t^{1}, \cdots \bs{{\theta}}_t^{j}]$ where all joint rotations except the root are replaced with simulated values (without $\widehat{\cdot}$). This amounts to setting the non-root joint goals to be identity when computing the goal states: $ \goalstatefailrec \triangleq (  \bs{\hat{\theta}}_t^{\prime} \ominus \bs{\theta}_t, \bs{\hat{p}}_t^{\prime} -  \bs{p}_t, \bs{\hat{v}}_t^{\prime} -  \bs{v}_t, \bs{\hat{\omega}}_t^{\prime} -  \bs{\omega}_t, \bs{\hat{\theta}}_t^{\prime}, \bs{\hat{p}}_t^{\prime})$. $\goalstatefailrec$ thus collapse from an imitation objective to a point-goal \cite{Won2022-jy} objective where the only information provided is the relative position and orientation of the target root. When the reference root is too far ($> 5m$), we normalize $\bs{\hat{p}}_t^{\prime} -  \bs{p}_t$ as $\frac{5 \times  (\bs{\hat{p}}_t^{\prime} -  \bs{p}_t)}{\|\bs{\hat{p}}_t^{\prime} -  \bs{p}_t\|_2}$ to clamp the goal position. Once the humanoid is close enough (\eg $<0.5m$ ), the goal will switch back to full-motion imitation: 
\begin{equation}
\label{eqn:fail_recovery}
    \bs{s}^{\text{g}}_t = \begin{cases} \bs{s}^{\text{g}}_t & \|\bs{\hat{p}}^0_t - \bs{p}^0_t \|_2 \leq 0.5 \\ 
                    \goalstatefailrec & \text{otherwise}. \\ 
                            \end{cases}
\end{equation}
To create fallen states, we follow ASE \cite{Peng2022-vr} and randomly drop the humanoid on the ground at the beginning of the episode. The faraway state can be created by initializing the humanoid 2 $\sim$ 5 meters from the reference motion. The reward for fail-state recovery consists of the AMP reward $r^{\text{amp}}_t$, point-goal reward $ r^{\text{g-point}}_t$, and energy penality $r^{\text{energy}}_t $, calculated by the reward function $\rewardfuncfailrec$:
\begin{equation}
    r^{\text{g-recover}}_t = \rewardfuncfailrec(\state, \refp) = 0.5 r^{\text{g-point}}_t + 0.5 r^{\text{amp}}_t + 0.1 r^{\text{energy}}_t, 
\end{equation}
The point-goal reward is formulated as $r^{\text{g-point}}_t=\left(d_{t-1}-d_t\right)$ where $d_t$ is the distance between the root reference and simulated root at the time step $t$ \cite{Won2022-jy}. For training $\prim^{(F)}$, we use a handpicked subset of the AMASS dataset named $\bs{Q}^{\text{loco}}$ where it contains mainly walking and running sequences. Learning using only $\bs{Q}^{\text{loco}}$ coaxes the discriminator $\disc$ and the AMP reward $r^{\text{amp}}_t$ to bias toward simple locomotion such as walking and running. We do not initialize a new value function and discriminator while training the primitives and continuously fine-tune the existing ones.

\paragraph{Multiplicative Control}
Once each primitive has been learned, we obtain $\{\prim^{(1)} \cdots \prim^{(K)}, \prim^{(F)}\}$, with each primitive capable of imitating a subset of the dataset $\motiondata$. In Progressive Networks \cite{Rusu2016-go}, task switching is performed manually. In motion imitation, however, the boundary between hard and easy sequences is blurred. Thus, we utilize Multiplicative Control Policy (MCP) \cite{Peng2019-kf} and train an additional composer $\bs{\mathcal{C}}$ to dynamically combine the learned primitives. Essentially, we use the pretrained primitives as a informed search space for the composer $\bs{\mathcal{C}}$, and $\bs{\mathcal{C}}$ only needs to select which primitives to activate for imitation. Specifically, our composer $\bs{\mathcal{C}} (\bs{w}^{1:K+1}_t | \state)$ consumes the same input as the primitives and outputs a weight vector $\bs{w}^{1:K+1}_t \in \reals^{k + 1}$ to activate the primitives. Combining our composer and primitives, we have the PHC's output distribution: 
\begin{equation}
\scriptsize
    \policy(\ba_t \mid \state)=\frac{1}{\bs{\mathcal{C}}(\state)} \prod_i^k \prim^{(i)} (\ba^{(i)}_t \mid \state)^{\bs{\mathcal{C}}(\state)}, \quad \bs{\mathcal{C}}(\state) \geq 0. 
\end{equation}
As each $\prim^{(k)}$ is an independent Gaussian, the action distribution:
\begin{equation}
\scriptsize
\mathcal{N}\left(\frac{1}{\sum_{l}^k \frac{\composer_i(\state)}{\sigma_l^j(\state)}} \sum_{i}^k \frac{\composer_i(\state)}{\sigma_i^j(\state)} \mu_i^j(\state), \sigma^j(\state)=\left(\sum_{i}^k \frac{\composer_i(\bs s_t)}{\sigma_i^j(\state)}\right)^{-1}\right), 
\end{equation}
where $\mu_i^j(\state)$ corresponds to the $\prim^{\text{(i)}}$'s  $j^{\text{th}}$ action dimension. Unlike a Mixture of Expert policies that only activates one at a time (top-1 MOE), MCP combines the actors' distribution and activates all actors at the same (similar to top-inf MOE). Unlike MCP, we progressively train our primitives and make the composer and actor share the same input space. Since primitives are independently trained for different harder sequences, we observe that the composite policy sees a significant boost in performance. During composer training, we interleave fail-state recovery training. The training process is described in Alg.\ref{alg:train} and Fig.\ref{fig:train}.

\subsection{Connecting with Motion Estimators}
\label{sec:real-time}
Our PHC is task-agnostic as it only requires the next time-step reference pose $\kinp$ or the keypoint $\kint$ for motion tracking. Thus, we can use any off-the-shelf video-based human pose estimator or generator compatible with the SMPL kinematic structure. For driving simulated avatars from videos, we employ HybrIK \cite{Li2020-vc} and MeTRAbs \cite{Sarandi2020-rj, Sarandi2022-jh}, both of which estimate in the metric space with the important distinction that HybrIK outputs joint rotation $\kinr$ while MeTRAbs only outputs 3D keypoints $\kint$. For language-based motion generation, we use the Motion Diffusion Model (MDM) \cite{Tevet2022-pr}. MDM generates disjoint motion sequences based on prompts, and we use our controller's recovery ability to achieve in-betweening.

\input{assets/main/tables/mocap_imitation}

\input{assets/main/tables/noisy_imitation}

\section{Experiments}

We evaluate and ablate our humanoid controller's ability to imitate high-quality MoCap sequences and noisy motion sequences estimated from videos in Sec.\ref{sec:imitation}. In Sec.\ref{sec:fail}, we test our controller's ability to recovery from fail-state. As motion is best in videos, we provide extensive qualitative results in the supplementary materials. All experiments are run three times and averaged. 

\paragraph{Baselines} We compare with the SOTA motion imitator UHC \cite{Luo2021-gu} and use the official implementation. We compare against UHC both \textit{with and without} residual force control. 

\paragraph{Implementation Details} We uses four primitives (including fail-state recovery) for all our evaluations. PHC can be trained on a single NVIDIA A100 GPU; it takes around a week to train all primitives and the composer. Once trained, the composite policy runs at $>30$ FPS. Physics simulation is carried out in NVIDIA's Isaac Gym \cite{Makoviychuk2021-sx}. The control policy is run at 30 Hz, while simulation runs at 60 Hz. For evaluation, we do not consider body shape variation and use the mean SMPL body shape.  

\paragraph{Datasets} PHC is trained on the training split of the AMASS \cite{Mahmood2019-ki} dataset. We follow UHC \cite{Luo2021-gu} and remove sequences that are noisy or involve interactions of human objects, resulting in 11313 high-quality training sequences and 140 test sequences. To evaluate our policy's ability to handle unseen MoCap sequences and noisy pose estimate from pose estimation methods, we use the popular H36M dataset \cite{Ionescu2014-fb}. From H36M, we derive two subsets \textit{H36M-Motion*} and \textit{H36M-Test-Video*}. H36M-Motion* contains 140 high-quality MoCap sequences from the entire H36M dataset. H36M-Test-Video* contains 160 sequences of noisy poses estimated from videos in the H36M test split (since SOTA pose estimation methods are trained on H36M's training split). * indicates the removal of sequences containing human-chair interaction.

\paragraph{Metrics} We use a series of pose-based and physics-based metrics to evaluate our motion imitation performance. We report the success rate ($\success$) as in UHC \cite{Luo2021-gu}, deeming imitation unsuccessful when, at \textit{any point} during imitation, the body joints are on average $> 0.5m$ from the reference motion. $\success$ measures whether the humanoid can track the reference motion without losing balance or significantly lags behind. We also report the root-relative mean per-joint position error (MPJPE) $\mpjpe$ and the global MPJPE $\gmpjpe$ (in mm), measuring our imitator's ability to imitate the reference motion both locally (root-relative) and globally. To show physical realism, we also compare acceleration $\acc$ (mm/frame$^2$) and velocity $\vel$ (mm/frame) difference between simulated and MoCap motion. All the baseline and our methods are physically simulated, so we do not report any foot sliding or penetration. 

\subsection{Motion Imitation}
\label{sec:imitation}
\paragraph{Motion Imitation on High-quality MoCap}
Table\ref{tab:mocap_imitation} reports our motion imitation result on the AMASS train, test, and H36M-Motion* dataset. Comparing with the baseline \textbf{with RFC}, our method outperforms it on almost all metrics across training and test datasets. On the training dataset, PHC has a better success rate while achieving better or similar MPJPE, showcasing its ability to better imitate sequences from the training split. On testing, PHC shows a high success rate on unseen MoCap sequences from both the AMASS and H36M data. Unseen motion poses additional challenges, as can be seen in the larger per-joint error. UHC trained without residual force performs poorly on the test set, showing that it lacks the ability to imitate unseen reference motion. Noticeably, it also has a much larger acceleration error because it uses high-frequency jitter to stay balanced. Compared to UHC, our controller has a low acceleration error even when facing unseen motion sequences, benefiting from the energy penalty and motion prior.  Surprisingly, our keypoint-based controller is on par and sometimes outperforms the rotation-based one. This validates that the keypoint-based motion imitator can be a simple and strong alternative to the rotation-based ones.

\paragraph{Motion Imitation on Noisy Input from Video}
We use off-the-shelf pose estimators HybrIK \cite{Li2020-vc} and MeTRAbs \cite{Sarandi2020-rj} to extract joint rotation (HybrIK) and keypoints (MeTRAbs) using images from the H36M test set. As a post-processing step, we apply a Gaussian filter to the extracted pose and keypoints. Both HyBrIK and MeTRAbs are per-frame models that do not use any temporal information.  Due to depth ambiguity, monocular global pose estimation is highly noisy \cite{Sarandi2020-rj} and suffers from severe depth-wise jitter, posing significant challenge to motion imitators. We find that MeTRAbs outputs better global root estimation $\kint^0$, so we use its $\kint^0$ combined with HybrIK's estimated joint rotation $\kinr$ (HybrIK + Metrabs (root)). In Table\ref{tab:imitation_noisy}, we report our controller and baseline's performance on imitating these noisy sequences. Similar to results on MoCap Imitation, PHC outperforms the baselines by a large margin and achieves a high success rate ($\sim 90\%$). This validates our hypothesis that PHC is robust to noisy motion and can be used to drive simulated avatars directly from videos. Similarly, we see that keypoint-based controller (ours-kp) outperforms rotation-based, which can be explained by 1) estimating 3D keypoint directly from images is an easier task than estimating joint rotations, so keypoints from MeTRABs are of higher quality than joint rotations from HybrIK; 2) our keypoint-based controller is more robust to noisy input as it has the freedom to use any joint configuration to try to match the keypoints.

\paragraph{Ablations}
Table\ref{tab:abla} shows our controller trained with various components disabled. We perform ablation on the noisy input from H36M-Test-Image* to better showcase the controller's ability to imitate noisy data. First, we study the performance of our controller before training to recover from fail-state. Comparing row 1 (R1) and R2, we can see that relaxed early termination (RET) allows our policy to better use the ankle and toes for balance. R2 vs R3 shows that using MCP directly without our progressive training process boosts the network performance due to its enlarged network capacity. However, using the PMCP pipeline significantly boosts robustness and imitation performance (R3 vs. R4). Comparing R4 and R5 shows that PMCP is effective in adding fail-state recovery capability \textbf{without} compromising motion imitation. Finally, R5 vs. R6 shows that our keypoint-based imitator can be on-par with rotation-based ones, offering a simpler formulation where only keypoints is needed. For additional ablation on MOE vs. MCP, number of primitives, please refer to the supplement. 

\paragraph{Real-time Simulated Avatars} We demonstrate our controller's ability to imitate pose estimates streamed in real-time from videos. Fig.\ref{fig:qual} shows a qualitative result on a live demonstration of using poses estimated from an office environment. To achieve this, we use our keypoint-based controller and MeTRAbs-estimated keypoints in a streaming fashion. The actor performs a series of motions, such as posing and jumping, and our controller can remain stable. Fig.\ref{fig:qual} also shows our controller's ability to imitate reference motion generated directly from a motion language model MDM \cite{Tevet2022-pr}. We provide extensive qualitative results in our supplementary materials for our real-time use cases. 

\input{assets/main/tables/ablation}

\subsection{Fail-state Recovery}
\label{sec:fail}
To evaluate our controller's ability to recover from fail-state, we measure whether our controller can successfully reach the reference motion within a certain time frame. We consider three scenarios: 1) fallen on the ground, 2) far away from reference motion, and 3) fallen and far from reference. We use a single clip of standing-still reference motion during this evaluation. We generate fallen-states by dropping the humanoid on the ground and applying random joint torques for 150 time steps. We create the far-state by initializing the humanoid 3 meters from the reference motion. Experiments are run randomly 1000 trials. From Tab.\ref{tab:failure_rec} we can see that both of our keypoint-based and rotation-based controllers can recover from fall state with high success rate ($>$ 90\%) even in the challenging scenario when the humanoid is both fallen and far away from the reference motion. For a more visual analysis of fail-state recovery, see our supplementary videos. 

\input{assets/main/tables/fail_recovery}

\section{Discussions}

\paragraph{Limitations}
While our purposed PHC can imitate human motion from MoCap and noisy input faithfully, it does not achieve a 100\% success rate on the training set. Upon inspection, we find that highly dynamic motions such as high jumping and back flipping are still challenging. Although we can train single-clip controller to \textbf{overfit} on these sequences (see the supplement), our full controller often fails to learn these sequences. We hypothesize that learning such highly dynamic clips (together with simpler motion) requires more planning and intent (\eg running up to a high jump), which is not conveyed in the single-frame pose target $\refpn$ for our controller. The training time is also long due to our progressive training procedure. Furthermore, to achieve better downstream tasks, the current disjoint process (where the video pose estimator is unaware of the physics simulation) may be insufficient; tighter integration with pose estimation \cite{Yuan2021-rl, Luo2022-ux} and language-based motion generation \cite{Yuan2022-re} is needed. 

\paragraph{Conclusion and Future Work} We introduce Perpetual Humanoid Controller, a general purpose physics-based motion imitator that achieves high quality motion imitation while being able to recover from fail-states. Our controller is robust to noisy estimated motion from video and can be used to perpetually simulate a real-time avatar without requiring reset. Future directions include 1) improving imitation capability and learning to imitate 100\% of the motion sequences of the training set; 2) incorporating terrain and scene awareness to enable human-object interaction; 3) tighter integration with downstream tasks such as pose estimation and motion generation, \etc.

\paragraph{Acknowledgements} We thank Zihui Lin for her help in making the plots in this paper. Zhengyi Luo is supported by the Meta AI Mentorship (AIM) program.

{\small
\bibliographystyle{ieee_fullname}
\bibliography{paperpile}
}

\clearpage
\appendix{   
    \hypersetup{linkcolor=black}
    \begin{Large}
        \textbf{Appendix}
    \end{Large}
    \etocdepthtag.toc{mtappendix}
    \etocsettagdepth{mtchapter}{none}
    \etocsettagdepth{mtappendix}{subsection}
    \newlength\tocrulewidth
    \setlength{\tocrulewidth}{1.5pt}
    \parindent=0em
    \etocsettocstyle{\vskip0.5\baselineskip}{}
    \tableofcontents
}

\section*{Appendices}
\input{appendix}

\end{document}

%% file: assets/macros.tex
\usepackage[dvipsnames]{xcolor}
\usepackage{amsfonts}
\usepackage{bm}
\usepackage{amsmath}
\usepackage{mathtools}
\usepackage{multirow}
\usepackage{booktabs}
\usepackage{caption}
\usepackage{enumitem}
\usepackage{wrapfig,lipsum,booktabs}
\usepackage{caption}
\usepackage{bbm}
\usepackage{multirow}
\usepackage{pifont}
\usepackage[linesnumbered,ruled,vlined]{algorithm2e}

\SetCommentSty{mycommfont}
\SetAlCapFnt{\small} 
\usepackage{etoc}
\SetAlgorithmName{Algo}{Algo}{}

\renewcommand{\paragraph}[1]{{\vspace{1mm}\noindent \bf #1}.}

\newcommand{\ba}{\bs{a}}

\newcommand{\reals}{\mathbb{R}}
\newcommand{\composer}{\bs{\mathcal{C}}}
\newcommand{\prim}{\bs{\mathcal{P}}}
\newcommand{\disc}{\bs{\mathcal{D}}}
\newcommand{\valuefunc}{\bs{\mathcal{V}}}

\newcommand{\policy}{{\pi_{\text{PHC}}}}
\newcommand{\rewardfunc}{\bs{\mathcal{R}}}
\newcommand{\rewardfuncimitation}{\rewardfunc^{\text{imitation}}}
\newcommand{\rewardfuncfailrec}{\rewardfunc^{\text{recover}}}

\newcommand{\refps}{{\bs{\hat{q}}_{1:T}}}
\newcommand{\refp}{{\bs{\hat{q}}_{t}}}

\newcommand{\reft}{{\bs{\hat{p}}_{t}}}
\newcommand{\refts}{{\bs{\hat{p}}_{1:T}}}
\newcommand{\refr}{{\bs{\hat{\theta}}_{t}}}
\newcommand{\refrs}{{\bs{\hat{\theta}}_{1:T}}}
\newcommand{\refv}{{\bs{\hat{\dot{q}}}_{t}}}
\newcommand{\refvs}{{\bs{\hat{\dot{q}}}_{1:T}}}
\newcommand{\refav}{{\bs{\hat{\omega}}_{t}}}
\newcommand{\reflv}{{\bs{\hat{v}}_{t}}}
\newcommand{\refpn}{{\bs{\hat{q}}_{t+1}}}
\newcommand{\reftn}{{\bs{\hat{p}}_{t+1}}}
\newcommand{\refrn}{{\bs{\hat{\theta}}_{t+1}}}

\newcommand{\kinp}{{\bs{\tilde{q}}_{t}}}
\newcommand{\kint}{{\bs{\tilde{p}}_{t}}}
\newcommand{\kinr}{{\bs{\tilde{\theta}}_{t}}}

\newcommand{\simp}{{\bs{{q}}_{t}}}

\newcommand{\simv}{{\bs{\dot{q}}_{t}}}

\newcommand{\simlv}{{\bs{v}_{t}}}

\newcommand{\simr}{{\bs{{\theta}}_{t}}}
\newcommand{\simt}{{\bs{{p}}_{t}}}
\newcommand{\goalstate}{{\bs{s}^{\text{g}}_t}}
\newcommand{\goalstaterot}{{\bs{s}^{\text{g-rot}}_t}}
\newcommand{\goalstatekp}{{\bs{s}^{\text{g-kp}}_t}}
\newcommand{\goalstatefailrec}{\bs{s}^{\text{g-Fail}}_t}
\newcommand{\selfstate}{{\bs{s}^{\text{p}}_t}}
\newcommand{\state}{{\bs{s}_t}}
\newcommand{\action}{{\bs{a}_t}}
\newcommand{\motiondata}{\bs{\hat{Q}}}
\newcommand{\locomotiondata}{\bs{Q}^{\text{loco}}}
\newcommand{\hardmotiondata}{\bs{\hat{Q}}_{\text{hard}}}

\newcommand{\mpjpe}{E_\text{mpjpe}}
\newcommand{\gmpjpe}{E_\text{g-mpjpe}}
\newcommand{\acc}{E_\text{acc}}
\newcommand{\vel}{E_\text{vel}}
\newcommand{\success}{\text{Succ}}

\newcommand{\bs}[1]{\boldsymbol{#1}}
\newcommand{\cmark}{\ding{51}}%
\newcommand{\xmark}{\ding{55}}%

%% file: assets/main/figures/teaser.tex
\twocolumn[{
\renewcommand\twocolumn[1][]{#1}%
\maketitle
\vspace{-0.4in}
\begin{center}
    \centering
    \includegraphics[width=1\textwidth]{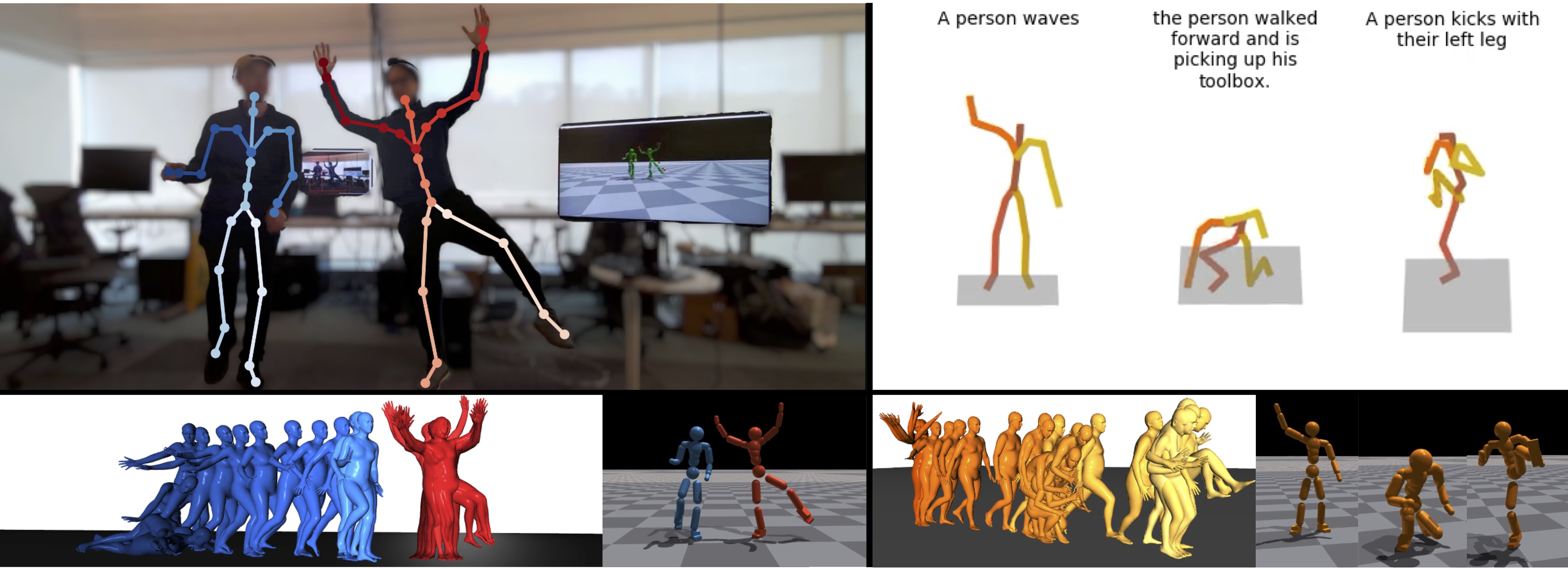}
    \vspace{-7mm}
    
    \captionof{figure}{\small{We propose a motion imitator that can naturally recover from falls and walk to far-away reference motion, perpetually controlling simulated avatars without requiring reset. Left: real-time avatars from video, where the blue humanoid recovers from a fall. Right: Imitating 3 \textit{disjoint} clips of motion generated from language, where our controller fills in the blank. The color gradient indicates the passage of time.  }}
    \label{fig:teaser}
\end{center}%
}]

%% file: assets/main/figures/train.tex
\begin{figure*}
\vspace{-1.5mm}
\begin{center}
\includegraphics[width=0.95\textwidth]{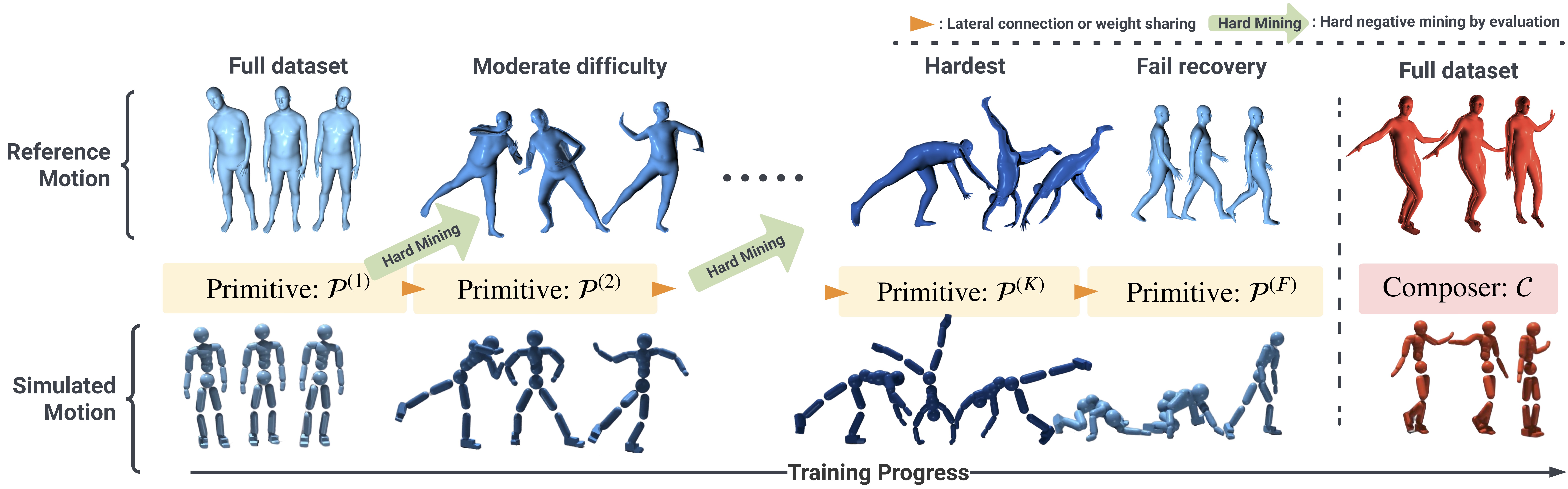}
\end{center}
\vspace{-5mm}
   \caption{\small{Our progressive training procedure to train primitives $\prim^{(1)}, \prim^{(2)}, \cdots, \prim^{(K)}$ by gradually learning harder and harder sequences. Fail recovery $\prim^{(F)}$ is trained in the end on simple locomotion data; a composer is then trained to combine these frozen primitives. } }
\vspace{-3.5mm}
\label{fig:train}
\end{figure*}

%% file: assets/main/figures/arch.tex
\begin{figure}
\vspace{-1.5mm}
\begin{center}
\includegraphics[width=0.5\textwidth]{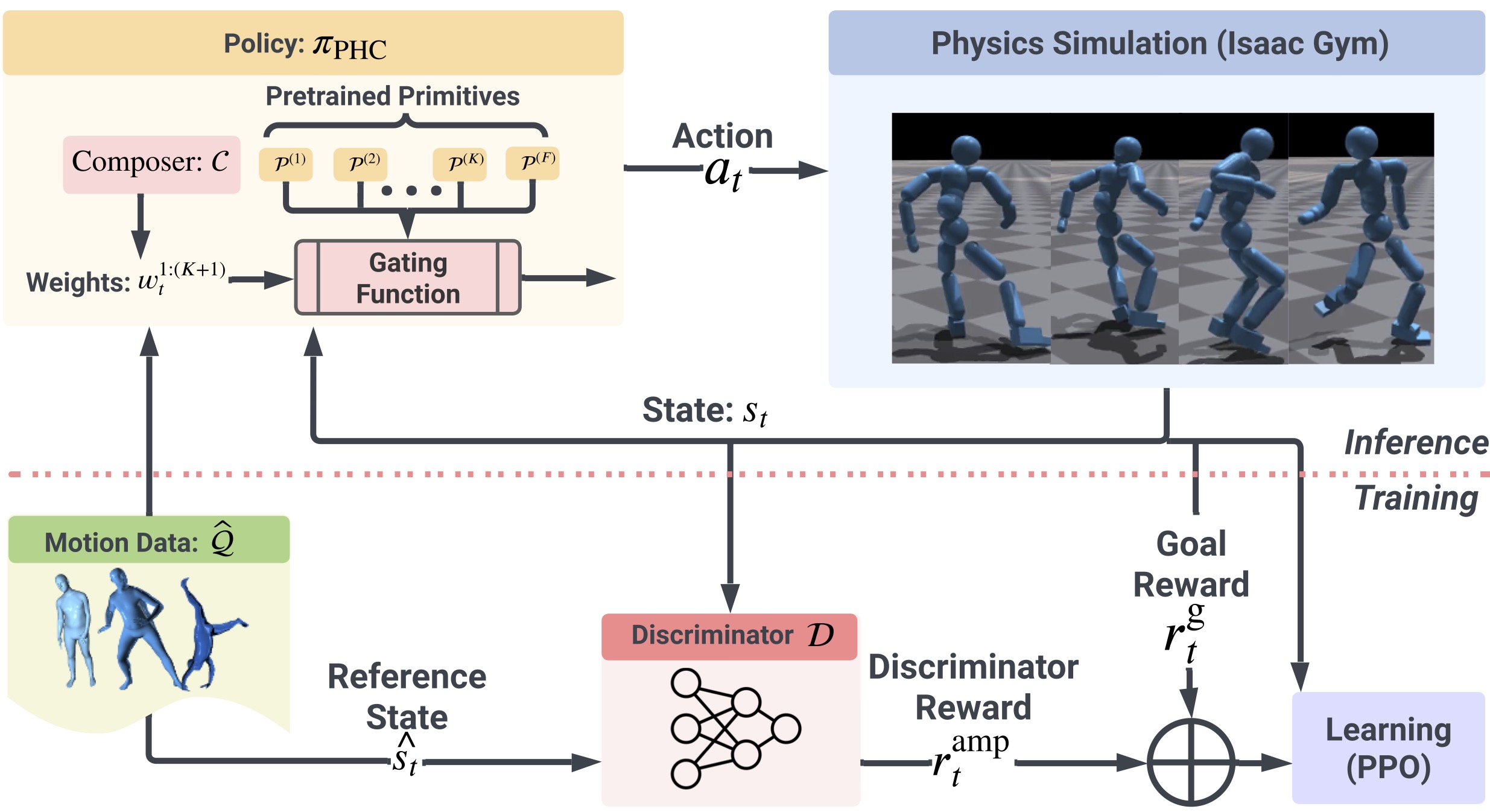}
\end{center}
\vspace{-5mm}
   \caption{\small{Goal-conditioned RL framework with Adversarial Motion Prior. Each primitive $\prim^{(k)}$ and composer $\composer$ is trained using the same procedure, and here we visualize the final product $\policy$.  }}
\vspace{-3.5mm}
\label{fig:arch}
\end{figure}

%% file: assets/main/tables/algo_train.tex
\begin{algorithm}[tb]
\scriptsize
  \caption{\small{Learn Progressive Multiplicative Control Policy}}\label{alg:train}
    \SetKwFunction{FMain}{TrainPPO}
     \SetKwProg{Fn}{Function}{:}{}
      \Fn{\FMain{$\pi$, $\motiondata^{(k)}$, $\disc$, $\valuefunc$, $\rewardfunc$}}{ 
        \While{not converged}{
        ${\bs M} \leftarrow \emptyset $ initialize sampling memory \,\;
            \While{${\bs M}$ not full}{
               $\refps \leftarrow$ sample motion from $\motiondata$ \,\;
                \For{$t \leftarrow 1...T$}{
                    $\bs s_t \leftarrow \left(\selfstate, \goalstate \right)$  \,\;
                    $\bs a_t \leftarrow {\pi}(\action | \state)$ \,\;
                     $\bs s_{t+1} \leftarrow \mathcal{T}(\bs s_{t+1} |\bs s_{t}, \bs a_t)$  \tcp*[f]{simulation}\,\;
                     $\bs r_t \leftarrow {\rewardfunc}(\state, \refpn)$  \,\;
                     store $(\bs s_{t}, \bs a_t, \bs r_t, \bs s_{t+1})$ into memory ${\bs M}$ \,\;
                }
            }
            $\prim^{(k)}, \valuefunc \leftarrow$ PPO update using experiences collected in ${\bs M}$ \,\;
            $\disc\leftarrow$ Discriminator update  using experiences collected in ${\bs M}$
            }
            \KwRet $\pi$ \;
      }

    \hrule
      
    \textbf{Input:}  Ground truth motion dataset $ \motiondata$\,\;

    $\disc$, $\valuefunc$, $ \hardmotiondata^{(1)} \leftarrow \motiondata$    \tcp*[f]{Initialize  discriminator,  value function, and dataset}\,\;
    
    \For{$k \leftarrow 1...K$}{
        Initialize $\prim^{(k)}$ \hspace{-2.2mm} \tcp*[f]{Lateral connection/weight sharing}\,\;
        $\prim^{(k)} \leftarrow$ \FMain{$\prim^{(k)}$, $\hardmotiondata^{(k+1)}$, $\disc$, $\valuefunc$, $\rewardfuncimitation$} \,\;
        $\hardmotiondata^{(k+1)} \leftarrow$ eval( $\prim^{(k)}$, $\motiondata^{(k)}$ )\,\;
        $\prim^{(k)} \leftarrow $ freeze $\prim^{(k)} $ \,\;
    }
    
    $\prim^{(F)} \leftarrow$ \FMain{$\prim^{(F)}$, $\locomotiondata$, $\disc$, $\valuefunc$, $\rewardfuncfailrec$} \tcp*[f]{Fail-state Recovery}\,\;

    $\policy \leftarrow {\{\prim^{(1)} \cdots \prim^{(K)}, \prim^{(F)}, \composer\}}$ \,\;
    $\policy  \leftarrow$ \FMain{$\policy$, $\motiondata$, $\disc$, $\valuefunc$, $\{\rewardfuncimitation, \rewardfuncfailrec\}$} \tcp*[f]{Train Composer}\,\;
\end{algorithm}

%% file: assets/main/figures/qual.tex
\begin{figure*}
\vspace{-1.5mm}
\begin{center}

\includegraphics[width=\textwidth]{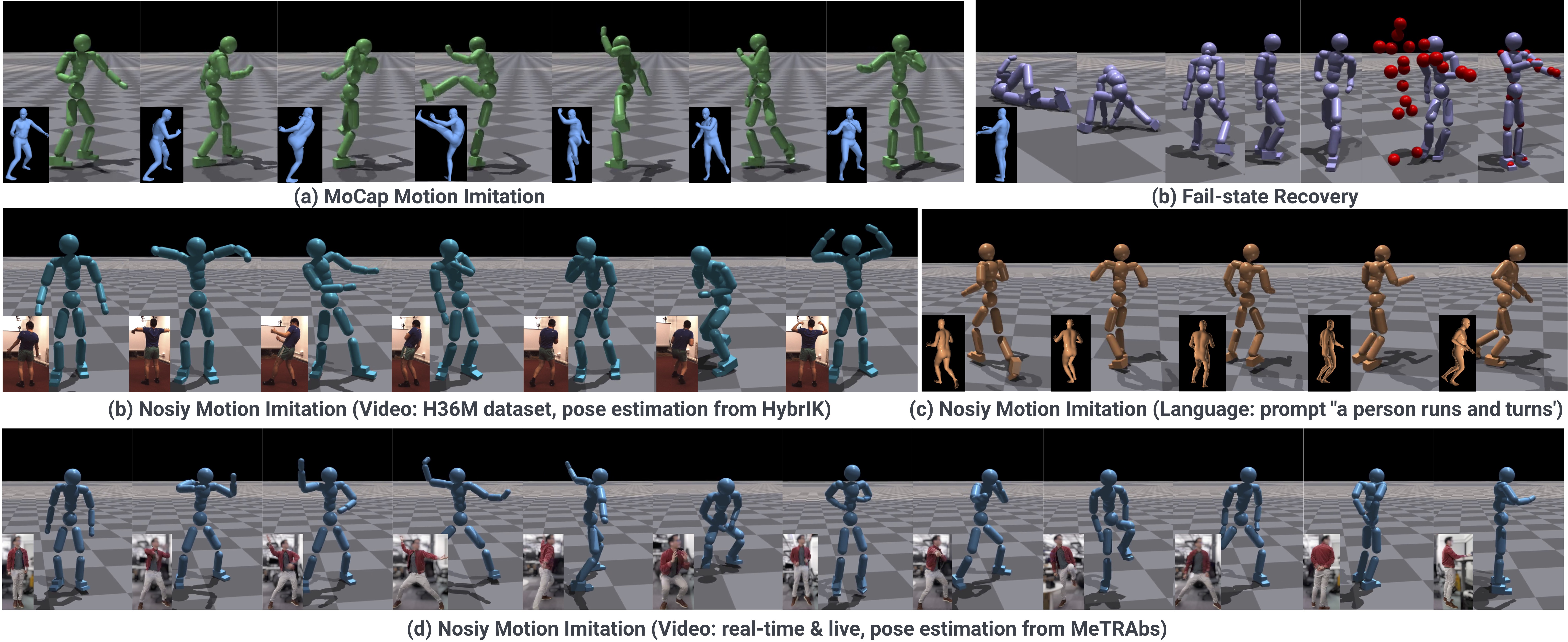}
\end{center}
\vspace{-5mm}
   \caption{\small{(a) Imitating high-quality MoCap -- spin and kick. (b) Recover from fallen state and go back to reference motion (indicated by red dots). (b) Imitating noisy motion estimated from video. (c) Imitating motion generated from language. (d) Using poses estimated from a webcam stream for a real-time simulated avatar.}}
\vspace{-3.5mm}
\label{fig:qual}
\end{figure*}

%% file: assets/main/tables/mocap_imitation.tex
\begin{table*}[t]
\caption{\small{Quantitative results on imitating MoCap motion sequences (* indicates removing sequences containing human-object interaction). AMASS-Train*, AMASS-Test*, and  H36M-Motion* contains 11313, 140, and 140 high-quality MoCap sequences, respectively. }
} \label{tab:mocap_imitation}
\centering
\resizebox{\linewidth}{!}{%
\begin{tabular}{lr|rrrrr|rrrrr|rrrrr}
\toprule
\multicolumn{2}{c}{} & \multicolumn{5}{c}{AMASS-Train* } & \multicolumn{5}{c}{AMASS-Test* } & \multicolumn{5}{c}{H36M-Motion*} 
\\ 
\midrule
Method  & $\text{RFC}$ & $\text{Succ} \uparrow$ & $E_\text{g-mpjpe}  \downarrow$ &  $E_\text{mpjpe} \downarrow $ &  $\text{E}_{\text{acc}} \downarrow$  & $\text{E}_{\text{vel}} \downarrow$ & $\text{Succ} \uparrow$ & $E_\text{g-mpjpe} \downarrow$ & $E_\text{mpjpe} \downarrow $ &  $\text{E}_{\text{acc}} \downarrow$  & $\text{E}_{\text{vel}} \downarrow$ & $\text{Succ} \uparrow$ & $E_\text{g-mpjpe} \downarrow$ &  $E_\text{mpjpe} \downarrow $ &  $\text{E}_{\text{acc}} \downarrow$  & $\text{E}_{\text{vel}} \downarrow$     \\ \midrule
\text{UHC} & \cmark &  {97.0 \%} & {36.4} & { 25.1}  & {4.4} & {5.9} & {96.4 \%} & {50.0} & { 31.2}  & {9.7} & {12.1} & {87.0\%} & {59.7} & {35.4} & {4.9} & {7.4}\\
\midrule

\text{UHC} & \xmark & {84.5 \%} & {62.7} & {39.6} & {10.9} & {10.9} & {62.6\%} & {58.2} & {98.1}  & {22.8}& {21.9} & {23.6\%} & {133.14} & {67.4} & {14.9} & {17.2}\\
\text{Ours} & \xmark & \textbf{98.9 \%} & \textbf{37.5} & \textbf{26.9} & \textbf{3.3} & \textbf{4.9} & {96.4\%} & \textbf{47.4} & \textbf{30.9} & {6.8} & \textbf{9.1} & {92.9\%} &{50.3} & \textbf{33.3} &{3.7}& \textbf{5.5}\\
\text{Ours-kp} & \xmark & {98.7\%} &  {40.7} & {32.3} & {3.5} & {5.5} & \textbf{97.1\%} & {53.1} & {39.5} & \textbf{7.5} & {10.4}   & \textbf{95.7\%} & \textbf{49.5} & {39.2} & \textbf{3.7} & {5.8} \\

\bottomrule 
\end{tabular}}
\end{table*}

%% file: assets/main/tables/noisy_imitation.tex
\begin{table}[t]
\caption{\small{Motion imitation on noisy motion. We use HybrIK\cite{Li2020-vc} to estimate the joint rotations $\kinr$ and uses MeTRAbs \cite{Sarandi2020-rj} for global 3D keypoints $\kint$}. HybrIK + MeTRAbs (root): using joint rotations $\kinr$ from HybrIK and root position $\kint^0$ from MeTRAbs. MeTRAbs (all keypoints): using all keypoints $\kint$ from MeTRAbs, only applicabile to our keypoint-based controller. } \label{tab:imitation_noisy}
\centering
\resizebox{\linewidth}{!}{%
\begin{tabular}{lrr|rrr}
\toprule
\multicolumn{3}{c}{} & \multicolumn{3}{c}{H36M-Test-Video*}  
\\ 
\midrule
Method  & $\text{RFC}$ & $\text{Pose Estimate}$ & $\text{Succ} \uparrow$ & $E_\text{g-mpjpe}  \downarrow$ &  $E_\text{mpjpe} \downarrow $   \\ \midrule
\text{UHC} & \cmark &  HybrIK + MeTRAbs (root) & {58.1\%} & {75.5} & {49.3}  \\
\midrule

\text{UHC} & \xmark & HybrIK + MeTRAbs (root) & {18.1\%} & {126.1} & {67.1} \\
\text{Ours} & \xmark & HybrIK + MeTRAbs (root) & {88.7\%} & \textbf{55.4} & \textbf{34.7} \\
\text{Ours-kp} & \xmark & HybrIK + MeTRAbs (root) & {90.0\%} & {55.8} & {41.0} \\
\text{Ours-kp} & \xmark & MeTRAbs (all keypoints) & \textbf{91.9\%} & {55.7} & 41.1 \\

\bottomrule 
\end{tabular}}
\end{table}

%% file: assets/main/tables/ablation.tex
\begin{table}[t]
\caption{\small{Ablation on components of our pipeline, performed using noisy pose estimate from HybrIK + Metrabs (root) on the H36M-Test-Video* data. RET: relaxed early termination. MCP: multiplicative control policy. PNN: progressive neural networks.}} \label{tab:abla}
\centering
\resizebox{\linewidth}{!}{%
\begin{tabular}{lrrrr|rrr}
\toprule
\multicolumn{2}{c}{} & \multicolumn{5}{c}{H36M-Test-Video*} 
\\ 
\midrule
 $\text{RET}$ & $\text{MCP}$  & $\text{PNN}$ &  $\text{Rotation}$ & $\text{Fail-Recover}$    & $\text{Succ} \uparrow$ & $E_\text{g-mpjpe}  \downarrow$ &    $E_\text{mpjpe} \downarrow $  \\ \midrule

  \xmark & \xmark & \xmark & \cmark & \xmark &  {51.2\%} & {56.2} & {34.4}  \\ %
  \cmark &\xmark & \xmark & \cmark & \xmark &  {59.4\%} & {60.2} & {37.2}  \\ %
  \cmark &\cmark & \xmark & \cmark & \xmark &  {66.2\%} & {59.0} & {38.3} \\  %
  \cmark &\cmark & \cmark & \cmark & \xmark &  {86.9\%} & \textbf{53.1} & \textbf{33.7} \\ %
  \midrule

  \cmark &\cmark & \cmark & \cmark & \cmark &  {88.7\%} & {55.4} & {34.7}  \\  %
  \cmark & \cmark & \cmark & \xmark & \cmark &  \textbf{90.0\%} & {55.8} & {41.0} \\  %
\bottomrule 
\end{tabular}}
\end{table}

%% file: assets/main/tables/fail_recovery.tex
\begin{table}[t]
\caption{\small{We measure whether our controller can recover from the fail-states by generating these scenarios (dropping the humanoid on the ground \& far from the reference motion) and measuring the time it takes to resume tracking. }}
\label{tab:failure_rec}
\centering
\resizebox{\linewidth}{!}{%
\begin{tabular}{l|rr|rr|rr}
\toprule
\multicolumn{1}{c|}{} & \multicolumn{2}{c|}{Fallen-State}  & \multicolumn{2}{c|}{Far-State}  & \multicolumn{2}{c}{Fallen + Far-State} 
\\ 
\midrule
 $\text{Method}$     & $\text{Succ-5s} \uparrow$ &  $\text{Succ-10s} \uparrow$ & $\text{Succ-5s} \uparrow$ &  $\text{Succ-10s} \uparrow$ & $\text{Succ-5s} \uparrow$ &  $\text{Succ-10s} \uparrow$    \\ \midrule
  \text{Ours}    &    { 95.0\% } & { 98.8\% } &{ 83.7\% } & { 99.5\% } & { 93.4\% } & { 98.8\% } \\
  \text{Ours-kp} &    { 92.5\% } & { 94.6\% } & { 95.1\% } & { 96.0\% } &{ 79.4\% } & { 93.2\% } \\
 
\bottomrule 
\end{tabular}}
\end{table}

%% file: appendix.tex
\section{Introduction}
In this document, we include additional details and results that are not included in the paper due to the page limit. In Sec.\ref{sec:impl-supp}, we include additional details for training, avatar use cases, and progressive neural networks (PNN) \cite{Rusu2016-go}. In Sec.\ref{sec:res-supp}, we include additional ablation results. Finally, in Sec.\ref{sec:discu}, we provide an extended discussion of limitations, failure cases, and future work. 

Extensive qualitative results are provided on the \href{https://zhengyiluo.github.io/PHC}{project page}. We highly encourage our readers to view them to better understand the capabilities of our method. Specifically, we show our method's ability to imitate high-quality MoCap data (both train and test) and noisy motion estimated from video. We also demonstrate real-time video-based (single- and multi-person) and language-based avatar (single- and multiple-clips) use cases. Lastly, we showcase our fail-state recovery ability.

\section{Implementation Details}
\label{sec:impl-supp}

\subsection{Training Details}
\paragraph{Humanoid Construction} Our humanoid can be constructed from any kinematic structure, and we use the SMPL humanoid structure as it has native support for different body shapes and is widely adopted in the pose estimation literature. Fig.\ref{fig:body} shows our humanoid constructed based on randomly selected gender and body shape from the AMASS dataset. The simulation result can then be exported and rendered as the SMPL mesh. We showcase two types of constructed humanoid: capsule-based and mesh-based. The capsule-based humanoid is constructed by treating body parts as simple geometric shapes (spheres, boxes, and capsules). The mesh-based humanoid is constructed following a procedure similar to SimPoE\cite{Yuan2021-rl}, where each body part is created by finding the convex hull of all vertices assigned to each bone. The capsule humanoid is easier to simulate and design, whereas the mesh humanoid provides a better approximation of the body shape to simulate more complex human-object interactions. We find that mesh-based and capsule-based humanoids do not have significant performance differences (see Sec.\ref{sec:res-supp}) and conduct all experiments using the capsule-based humanoid. For a fair comparison with the baselines, we use the mean body shape of the SMPL with neutral gender for all evaluations and show qualitative results for shape variation. For both types of humanoids, we scale the density of geometric shapes so that the body has the correct weight (on average 70 kg). All inter-joint collisions are enabled for all joint pairs except for between parent and child joints. Collision between humanoids can be enabled and disabled at will (for multi-person use cases). 

\input{assets/supp/figures/body}

\paragraph{Training Process}
During training, we randomly sample motion from the current training set $\motiondata^{(k)}$ and normalize it with respect to the simulated body shape by performing forward kinematics using $\refrs$. Similar to UHC \cite{Luo2021-gu}, we adjust the height of the root translation $\reft^0$ to make sure that each of the humanoid's feet touches the ground at the beginning of the episode. We use parallelly simulate 1536 humanoids for training all of our primitives and composers. Training takes around 7 days to collect approximately 10 billion samples. When training with different body shapes, we randomly sample valid human body shapes from the AMASS dataset and construct humanoids from them. Hyperparamters used during training can be found in Table.\ref{tab:supp_hyper}

\paragraph{Data Preparation} We follow similar procedure to UHC \cite{Luo2021-gu} to filter out AMASS sequences containing human object interactions. We remove all sequences that sits on chairs, move on treadmills, leans on tables, steps on stairs, floating in the air \etc, resulting in 11313 high-quality motion sequences for training and 140 sequences for testing. We use a heuristic-based filtering process based on \ie identifying the body joint configurations corresponding to the sitting motion or counting number of consecutive airborne frames.

\paragraph{Runtime}
Once trained, our PHC can run in real time ($\sim 32$FPS) together with simulation and rendering, and around ($\sim 50$FPS) when run without rendering. Table.\ref{tab:supp-abla} shows the runtime of our method with respect to the number of primitives, architecture, and humanoid type used. 

\paragraph{Model Size}
The final model size (with four primitives) is 28.8 MB, comparable to the model size of UHC (30.4 MB). 

\input{assets/supp/tables/hyperparmaters}

\subsection{Real-time Use Cases}

\paragraph{Real-time Physics-based Virtual Avatars from Video} To achieve real-time physics-based avatars driven by video, we first use Yolov8\cite{Jocher_Glenn_and_Chaurasia_Ayush_and_Qiu_Jing_undated-sf} for person detection. For pose estimation, we use MeTRAbS \cite{Sarandi2020-rj} and HybrIK \cite{Li2020-vc} to provide 3D keypoints $\kint$ and rotation $\kinr$. MeTRAbs is a 3D keypoint estimator that computes 3D joint positions $\kint$ in the absolute global space (rather than in the relative root space). HybrIK is a recent method for human mesh recovery and computes joint angles $\kinr$ and root position $\kint^{0}$ for the SMPL human body. One can recover the 3D keypoints $\kint$ from joint angles $\kinr$ and root position $\kint^{0}$ using forward kinematics. Both of these methods are causal, do not use any temporal information, and can run in real-time ($\sim 30$FPS). Estimating 3D keypoint location from image pixels is an easier task than regressing joint angles, as 3D keypoints can be better associated with features learned from pixels. Thus, both HybrIK and MeTRAbs estimate 3D keypoints $\kint$, with HybrIK containing an additional step of performing learned inverse kinematics to recover joint angles $\kinr$. We show results using both of these off-the-shelf pose estimation methods, using MeTRAbs with our keypoint-based controller and HybrIK with our rotation-based controller. Empirically, we find that MeTRAbs estimates more stable and accurate 3D keypoints, potentially due to its keypoint-only formulation. We also present a real-time \textbf{multi-person} physics-based human-to-human interaction use case, where we drive multiple avatars and enable inter-humanoid collision. To support multi-person pose estimation, we use OCSort \cite{Cao2022-ya} to track individual tracklets and associate poses with each person. Notice that real-time use cases pose additional challenges than offline processing: detection, pose/keypoint estimation, and simulation all need to run at real-time at around 30 FPS, and small fluctuations in framerate could lead to unstable imitation and simulation. To smooth out noisy depth estimates, we use a Gaussian filter to smooth out estimates from t-120 to t, and use the ``mirror" setting for padding at boundary.

\paragraph{Virtual Avatars from Language} For language-based motion generation, we adopt MDM \cite{Tevet2022-pr} as our text-to-motion model. We use the official implementation, which generates 3D keypoints $\kint$ by default and connects it to our keypoint-based imitator. MDM generates fixed-length motion clips, so additional blending is needed to combine multiple clips of generated motion. However, since PHC can naturally go to far-away reference motion and handles disjoint between motion clips, we can naively chain together multiple clips of motion generated by MDM and create coherent and physically valid motion from multiple text prompts. This enables us to create a simulated avatar that can be driven by a continuous stream of text prompts.

\subsection{Progressive Neural Network (PNN) Details}  A PNN \cite{Rusu2016-go} starts with a single primitive network $\prim^{(1)}$ trained on the full dataset $\motiondata$. Once $\prim^{(1)}$ is trained to convergence on the entire motion dataset $\motiondata$ using the imitation task, we create a subset of hard motions by evaluating $\prim^{(1)}$ on $\motiondata$. Sequences that $\prim^{(1)}$ fails forms $\hardmotiondata^{(1)}$. We then freeze the parameters of $\prim^{(1)}$ and  create a new primitive $\prim^{(2)}$ (randomly initialized) along with lateral connections that connect each layer of $\prim^{(1)}$ to $\prim^{(2)}$. Given the layer weight $\bs{W}_i^{(k)}$, activation function $f$, and the learnable lateral connection weights $\bs{U}_i^{(j:k)}$, we have the hidden activation $\bs{h}_i^{(k)}$ of the $i^{\text{th}}$ layer of $k^{\text{th}}$ primitive as:

\input{assets/supp/figures/pnn}

\small
\begin{equation}
    \bs{h}_i^{(k)}=f\left(\bs{W}_i^{(k)} \bs{h}_{i-1}^{(k)}+\sum_{j<k} \bs{U}_i^{(j:k)} \bs{h}_{i-1}^{(j)}\right).
\end{equation}
Fig.\ref{fig:sup-pnn} visualizes the PNN with the lateral connection architecture. Essentially, except for the first layer, each subsequent layer receives the activation of the previous layer processed by the learnable connection matrices $\bs{U}_i^{(j:k)}$. We do not use any adapter layer as in the original paper. As an alternative to lateral connection, we explore weight sharing and warm-starts the primitive with the weights from the previous one (as opposed to randomly initialized). We find both methods equally effective (see Sec.\ref{sec:res-supp}) when trained with the same hard-negative mining procedure, as each newly learned primitive adds new sequences that PHC can imitate. The weight sharing strategy significantly decreases training time as the policy starts learning harder sequences with basic motor skills. We use weight sharing in all our main experiments.

\section{Supplementary Results}
\label{sec:res-supp}

\input{assets/supp/figures/forget}
\subsection{Categorizing the Forgetting Problem}
\label{sec:res-supp-forget}
As mentioned in the main paper, one of the main issues in learning to mimic a large motion dataset is the forgetting problem. The policy will learn new sequences while forgetting the ones already learned. In Fig.\ref{fig:sup-forget}, we visualize the sequences that the policy fails to imitate during training. Starting from the 12.5k epoch, each evaluation shows that some sequences are learned, but the policy will fail on some already learned sequences. The staircase pattern indicates that when learning sequences failed previously, the policy forgets already learned sequences. Numerically, each evaluation has around 30\% overlap of failed sequences (right end side). The 30\% overlap contains the backflips, cartwheeling, and acrobatics; motions that the policy consistently fails to learn when trained together with other sequences. We hypothesize that these remaining sequences (around 40) may require additional sequence-level information for the policy to learn properly together with other sequences. 

\textbf{Fail-state recovery} Learning the fail-state recovery task can also lead to forgetting previously learned imitation skills. To verify this, we evaluate $\prim^{(F)}$ on the H36M-Test-Video dataset, which leads to a performance of Succ: 42.5\%, $E_\text{g-mpjpe}$: 87.3, and $E_\text{mpjpe}$: 55.9, which is much lower than the single primitive $\prim^{(1)}$ performance of Succ: 59.4\%, $E_\text{g-mpjpe}$: 60.2, and $E_\text{mpjpe}$: 34.4. Thus, learning the fail-state recovery task may lead to severe forgetting of the imitation task, motivating our PMCP framework to learn separate primitives for imitation and fail-state recovery.

\subsection{Additional Ablations}

In this section, we provide additional ablations of the components of our framework. Specifically, we study the effect of MOE vs. MCP, lateral connection vs. weight sharing, and the number of primitives used. We also report the inference speed (counting network inference and simulation time). All experiments are carried out with the rotation-based imitator and incorporate the fail state recovery primitive $\prim^{(F)}$ as the last primitive. 

\paragraph{PNN Lateral Connection vs. Weight Sharing} As can be seen in Table \ref{tab:supp-abla}, comparing Row 1 (R1) and R7 , we can see that PNN with lateral connection and weight sharing produce similar performance, both in terms of motion imitation and inference speed. This shows that \textit{in our setup}, the weight sharing scheme is an effective alternative to lateral connections. This can be explained by the fact that in our case,  each ``task" on which the primitives are trained is similar and does not require lateral connection to choose whether to utilize prior experiences or not. 

\paragraph{MOE vs. MCP} The difference between the top-1 mixture of experts (MOE) and multiplicative control (MCP) is discussed in detail in the MCP paper \cite{Peng2019-kf}: top-1 MOE only activates one expert at a time, while MCP can activate all primitives at the same time. Comparing R2 and R7, as expected, we can see that top-1 MOE is slightly inferior to MCP. Since all of our primitives are pretrained and frozen, theoretically a perfect composer should be able to choose the best primitive based on input for both MCP and MOE. MCP, compared to MOE, can activate all primitives at once and search a large action space where multiple primitives can be combined. Thus, MCP provides better performance, while MOE is not far behind. This is also observed by CoMic\cite{Hasenclever_undated-cs}, where they observe similar performance between mixture and product distributions when used to combine subnetworks. Note that top-inf MOE is similar to MCP where all primitives can be activated. 

\input{assets/supp/tables/supp_ablation}

\paragraph{Capsule vs. Mesh Humanoid} Comparing R3 and R7, we can see that mesh-based humanoid yield similar performance to capsule-based ones. It does slow down simulation by a small amount (30 FPS vs. 32 FPS), as simulating mesh is more compute-intensive than simulating simple geometries like capsules. 

\paragraph{Number of primitives} Comparing R4, R5, R6, R7, and R8, we can see that the performance increases as the number of primitives increases. Since the last primitive $\prim^{(F)}$ is for fail-state recovery and does not provide motion imitation improvement, R5 is similar to the performance of models trained without PMCP (R4). As the number of primitives grows from 2 to 3, we can see that the model performance grows quickly, showing that MCP is effective in combining pretrained primitives to achieve motion imitation. Since we are using relatively small networks, the inference speed does not change significantly with the number of primitives used. We notice that as the number of primitives grows, $\motiondata^{(k)}$ becomes more and more challenging. For instance, $\motiondata^{(4)}$ contains mainly highly dynamic motions such as high-jumping, back flipping, and cartwheeling, which are increasingly difficult to learn together. We show that (see supplementary webpage) we can overfit these sequences by training on them only, yet it is significantly more challenging to learn them together. Motions that are highly dynamic require very specific steps to perform (such as moving while airborne to prepare for landing). Thus, the experiences collected when learning these sequences together may contradict each other: for example, a high jump may require a high speed running up, while a cartwheel may require a different setup of foot-movement. A per-frame policy that does not have sequence-level information may find it difficult to learn these sequences together. Thus, sequence-level or information about the future may be required to learn these high dynamic motions together. In general, we find that using 4 primitives is most effective in terms of training time and performance, so for our main evaluation and visualizations, we use \textbf{4-primitive models}.

\section{Extended Limitation and Discussions}
\label{sec:discu}

\paragraph{Limitation and Failure Cases} As discussed in the main paper, PHC has yet to achieve 100\% success rate on the AMASS training set. With a 98.9\% success rate, PHC can imitate \textit{most} of our daily motion without losing balance, but can still struggle to perform more dynamic motions, such as backflipping. For our real-time avatar use cases, we can see a noticeable degradation in performance from the offline counterparts. This is due to the following:
\begin{itemize}
    \item Discontinuity and noise in reference motion. The inherent ambiguity in monocular depth estimation can result in noisy and jittery 3D keypoints, particularly in the depth dimension. These small errors, though sometimes imperceptible to the human eye,  may provide PHC with incorrect movement signals, leaving insufficient time for appropriate reactions.  Velocity estimation is also especially challenging in real-time use cases, and PHC relies on stable velocity estimation to infer movement cues.  
    \item Mismatched framerate. Since our PHC assumes 30 FPS motion input, it is essential for pose estimates from video to match for a more stable imitation. However, few pose estimators are designed to perform real-time pose estimation ($\geq$ 30 FPS), and the estimation framerate can fluctuate due to external reasons, such as the load balance on computers. 
    \item For multi-person use case, tracking and identity switch can still happen, leading to a jarring experience where the humanoids need to switch places. 
\end{itemize}
A deeper integration between the pose estimator and our controller is needed to further improve our real-time use cases. As we do not explicitly account for camera pose, we assume that the webcam is level with the ground and does not contain any pitch or roll. Camera height is manually adjusted at the beginning of the session. The pose of the camera can be taken into account in the pose estimation stage. Another area of improvement is naturalness during fail-state recovery. While our controller can recover from fail-state in a human-like fashion and walks back to resume imitation, the speed and naturalness could be further improved. Walking gait, speed, and tempo during fail-state recovery exhibits noticeable artifacts, such as asymmetric motion, a known artifact in AMP \cite{Peng2021-xu}. During the transition between fail-state recovery and motion imitation, the humanoid can suddenly jolt and snap into motion imitation. Further investigation (\eg better reward than the point-goal formulation, additional observation about trajectory) is needed.

\paragraph{Discussion and Future Work} We propose the perpetual humanoid controller, a humanoid motion imitator capable of imitating large corpus of motion with high fidelity. Paired with its ability to recover from fail-state and go back to motion imitation, PHC is ideal for simulated avatar use cases where we no longer require reset during unexpected events. We pair PHC with a real-time pose estimator to show that it can be used in a video-based avatar use case, where the simulated avatar imitates motion performed by the actors perpetually without requiring reset. This can empower future virtual telepresence and remote work, where we can enable physically realistic human-to-human interactions. We also connect PHC to a language-based motion generator to demonstrate its ability to mimic generated motion from text. PHC can imitate multiple clips by performing motion inbetweening. Equipped with this ability, future work in embodied agents can be paired with a natural language processor to perform complex tasks. Our proposed PMCP can be used as a general framework to enable progressive RL and multi-task learning. In addition, we show that one can use \textbf{only 3D keypoint} as motion input for imitation, alleviating the requirement of estimating joint rotations. Essentially, we use PHC to perform inverse kinematics based on the input 3D keypoints and leverages the laws of physics to regulate its output. We believe that PHC can also be used in other areas such as embodied agents and grounding, where it can serve as a low-level controller for high-level reasoning functions.

%% file: assets/supp/figures/body.tex
 \begin{figure}
\vspace{-1.5mm}
\begin{center}
\includegraphics[width=0.5\textwidth]{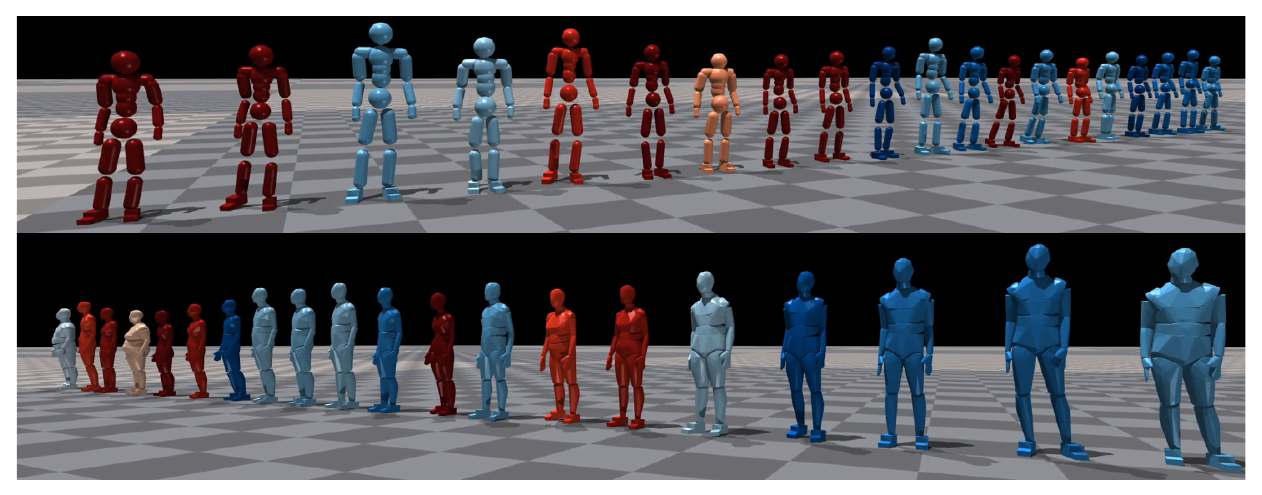}
\end{center}
\vspace{-5mm}
   \caption{\small{Our framework can support body shape and gender variations. Here we showcase humanoids of different gender and body proportion holding a standing pose. We construct two kinda of humanoids: capsule-based (top) and mesh-based (bottom). Red: female, Blue: male. Color gradient indicates weight. }}
\label{fig:body}
\end{figure}

%% file: assets/supp/tables/hyperparmaters.tex
\begin{table}[t]
\caption{Hyperparameters for PHC. $\sigma$: fixed variance for policy. $\gamma$: discount factor. $\epsilon$: clip range for PPO } 
\label{tab:supp_hyper}
\centering
\resizebox{\linewidth}{!}{%
\begin{tabular}{lccccccc}
\toprule
   & Batch Size & Learning Rate  &  $\sigma$  & $\gamma$ & $\epsilon$
  \\ \midrule
Value     & 1536 & $5 \times 2^{-5}$ &  0.05  & 0.99 & 0.2 & 
\\ \midrule
&$w_{\text{jp}}$ & $w_{\text{jr}}$ &$w_{\text{jv}}$ & $w_{\text{j}\omega}$ 
  \\ \midrule
Value   & 0.5    & 0.3   & 0.1   & 0.1    &    & \\
\bottomrule 
\end{tabular}}\\ 
\end{table}

%% file: assets/supp/figures/pnn.tex
 \begin{figure}
\vspace{-1.5mm}
\begin{center}
\includegraphics[width=0.5\textwidth]{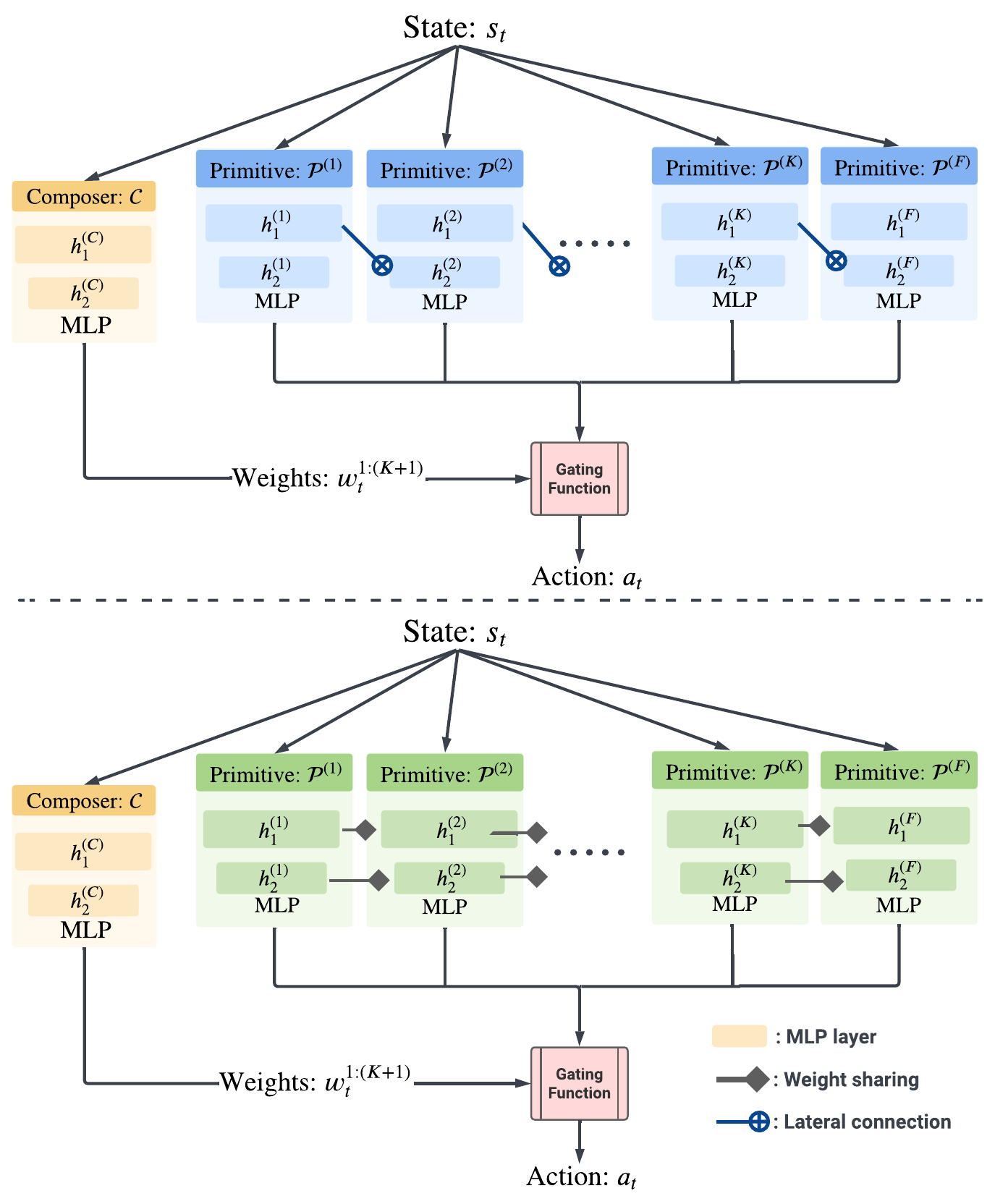}
\end{center}
\vspace{-5mm}
   \caption{\small{Progressive neural network architecture. Top: PNN with lateral connection. Bottom: PNN with weight sharing. $h_i^{(j)}$ indicates hidden activation of $j^{\text{th}}$ primitive's $i^{\text{th}}$ layer.}}
\vspace{-3.5mm}
\label{fig:sup-pnn}
\end{figure}

%% file: assets/supp/figures/forget.tex
 \begin{figure}
\vspace{-1.5mm}
\begin{center}
\includegraphics[width=0.45\textwidth]{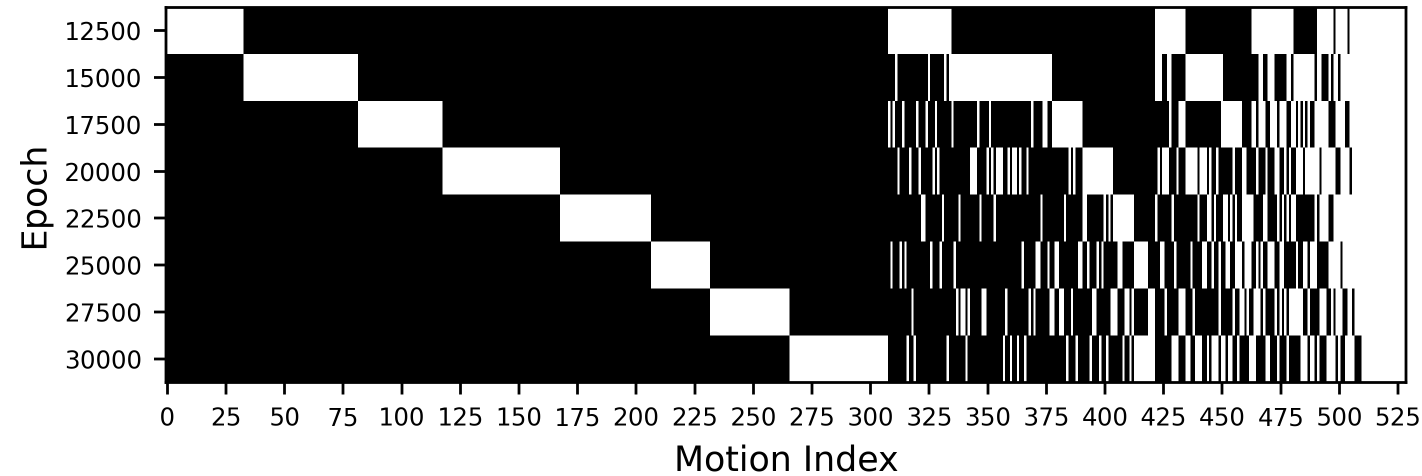}
\end{center}
\vspace{-5mm}
   \caption{\small{Here we plot the motion indexes that the policy fails on over training time; we only plot the 529 sequences that the policy has failed on over these training epoches. A white
pixel denotes that sequence is can be successfully imitated at the given epoch, and a black pixel denotes an unsuccessful imitation. We can see that while there are 30 sequences that the policy consistently fails on, the remaining can be learned and then forgotten as training progresses. The staircase pattern indicates that the policy fails on new sequences each time it learns new ones.}}
\vspace{-3.5mm}
\label{fig:sup-forget}
\end{figure}

%% file: assets/supp/tables/supp_ablation.tex
\begin{table}[t]
\caption{\small{Supplmentary ablation on components of our pipeline, performed using noisy pose estimate from HybrIK + Metrabs (root) on the H36M-Test-Video* data. MOE: top-1 mixture of experts. MCP: multiplicative control policy. PNN: progressive neural networks. Type: between Cap (capsule) and mesh-based humanoids. All models are trained with the same procedure.}} \label{tab:supp-abla}
\centering
\resizebox{\linewidth}{!}{%
\begin{tabular}{rrrrrr|rrrr}
\toprule
\multicolumn{5}{c}{} & \multicolumn{4}{c}{H36M-Test-Video*} 
\\ 
\midrule
  $\text{PNN-Lateral}$ & $\text{PNN-Weight}$ & $\text{MOE}$ & $\text{MCP}$ & $\text{Type}$ & $\text{\# Prim}$   & $\text{Succ} \uparrow$ & $E_\text{g-mpjpe}  \downarrow$ &    $E_\text{mpjpe} \downarrow $ &$\text{FPS}$  \\ \midrule

  \cmark &\xmark & \xmark & \xmark & Cap & 4 &  {87.5\%} & {55.7} & {36.2} & {32}\\  %
  \xmark &\cmark & \cmark & \xmark & Cap & 4 &  {87.5\%} & {56.3} & {34.3} & {33} \\ %
  \xmark &\cmark & \xmark & \cmark & Mesh & 4 &  {86.9\%} & {62.6} & {39.5} & {30} \\\midrule %
  \xmark &\xmark & \xmark & \xmark & Cap & 1 &  {59.4\%} & {60.2} & {37.2} & {32}\\  %
  \xmark &\cmark & \xmark & \cmark & Cap & 2 &  {65.6\%} & {58.7} & {37.3} & {32}\\  %
  \xmark &\cmark & \xmark & \cmark & Cap & 3 &  {80.9\%} & {56.8} & {36.1} & {32}\\ %
  \xmark &\cmark & \xmark & \cmark & Cap & 4 &  \textbf{88.7\%} & \textbf{55.4} & \textbf{34.7} & {32} \\  %
  \xmark &\cmark & \xmark & \cmark & Cap & 5 &  {87.5\%} & {57.7} & {36.0} & {32}\\ %

\bottomrule 
\end{tabular}}
\end{table}